\documentclass[twocolumn,secnumarabic,amssymb, nobibnotes, aps, prd]{revtex4-1}
\usepackage{times,graphicx,times,amsthm,amsfonts,amsmath,amscd,amssymb}
\usepackage{algpseudocode}

\newcommand{\cG}{{\mathcal G}}
\newcommand{\cE}{{\mathcal E}}
\newcommand{\cC}{{\mathcal C}}

\newtheorem{dfn}{Definition}

\newtheorem{thm}{Theorem}
\newtheorem{lemma}{Lemma}

\newtheorem{prop}{Proposition}
\newcommand{\paranth}[1]{\left(#1\right)}

\newcommand{\curly}[1]{\left\{#1\right\}}
\newcommand{\scg}{sequential community graph }

\newcommand{\scgs}{sequential community graphs }

\begin{document}

\title{Leaders, Followers, and Community Detection}%

\author{Dhruv Parthasarathy}%
\affiliation{Department of Electrical Engineering and Computer Science, MIT, Cambridge, MA 02139, USA}
\author{Devavrat Shah}%
\affiliation{Department of Electrical Engineering and Computer Science, MIT, Cambridge, MA 02139, USA}
\author{Tauhid Zaman}%
\affiliation{Sloan School of Management, MIT, Cambridge, MA 02139, USA}

\begin{abstract}

Communities in social networks or graphs are sets of well-connected, overlapping vertices. The effectiveness of a community detection algorithm is determined by accuracy in finding the ground-truth communities and ability to scale with the size of the data. In this work, we provide three contributions. First, we show that a popular measure of accuracy known as the F1 score, which is between 0 and 1, with 1 being perfect detection, has an “information lower bound” is 0.5. We provide a trivial algorithm that produces communities with an F1 score of 0.5 for any graph!  Somewhat surprisingly, we find that popular algorithms such as modularity optimization, BigClam and CESNA have F1 scores less than 0.5 for the popular IMDB graph. To rectify this, as the second contribution we propose a generative model for community formation, the sequential community graph, which is motivated by the formation of social networks. Third, motivated by our generative model, we propose the “leader-follower algorithm” (LFA). We prove that it recovers all communities for sequential community graphs by establishing a structural result that sequential community graphs are chordal. For a large number of popular social networks, it recovers communities with a much higher F1 score than other popular algorithms. For the IMDB graph, it obtains an F1 score of 0.81. We also propose a modification to the LFA called the fast leader-follower algorithm (FLFA) which in addition to being highly accurate, is also fast, with a scaling that is almost linear in the graph / network size.

\end{abstract}
\maketitle

\section{Introduction}\label{sec:introduction}
%
\begin{figure}[t]
	\centering
	\includegraphics[scale = .25]{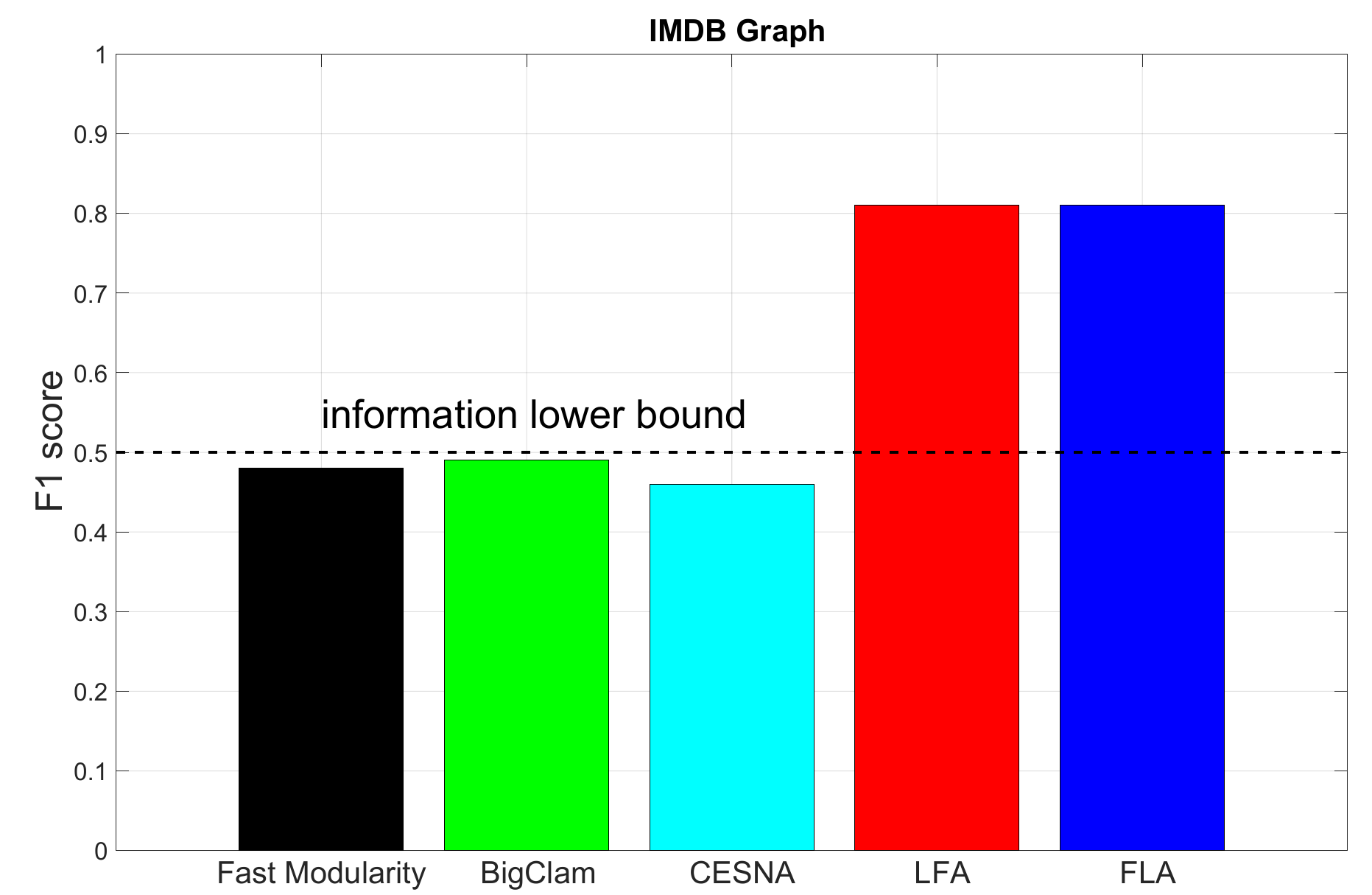}
	\caption{F1 score on IMDB graph for several community detection algorithms, including our leader-follower and fast leader-follower algorithms (LFA and FLFA). A lower bound of $0.5$ based on an algorithm that does not utilize {\em any information} is indicated by the dashed line.}  
	\label{fig:IMDB}
\end{figure}
Understanding community structure is an important and well studied problem in the analysis of social networks.
Communities represent a latent structure that is manifested through densely connected vertices.  For example,
a latent social group such co-workers may show up as a set of people in a social network connected by a 
dense set of edges.  While many community detection algorithms have been proposed (cf. \cite{fortunato2010community}), 
an important question is how to evaluate their performance.  One approach is to compare
the detected communities to a ground-truth set of communities if feasible.  In this case,
one needs to define some notion of \emph{distance} between two sets of communities.

In \cite{yangcommunity} the F1 score, which is based on concepts from information retrieval,
is used to assess the accuracy of community detection methods.  The score assigns a value between $0$
and $1$. It  gives a higher value to communities which are closer to the ground-truth communities. The
question is, what is a good F1 score? Clearly, 1 is an excellent score because it means perfect 
identification. But, for example, consider a popular IMDB graph for which we evaluate
three excellent algorithms from the literature: modularity optimization \cite{newman2006modularity}, 
CESNA \cite{yangcommunity} and BigClam \cite{yang2013overlapping}. Their respective
F1 scores for the IMDB graph of $0.48, 0.46$ and $0.49$ are shown in Figure \ref{fig:IMDB}. Are these scores 
{\em good}, {\em okay} or {\em terrible}?

\smallskip
\noindent
{\bf Our contributions. }  To answer this question, as an important contribution we establish 
a non-trivial lower bound for the F1 score. Specifically, we show that there exists a simple algorithm 
that can produce communities with an F1 score of $0.5$ for {\em any} graph {\em without} accessing 
the graph structure. That is, $0.5$ is {\em information lower bound} on community detection. 
In that sense, for the IMDB graph mentioned above, the F1 scores of modularity optimization, 
CESNA and BigClam are simply {\em terrible}: based on the F1 score, these algorithms are unable to extract {\em any meaningful information} from the graph structure. 

This clearly suggests that we need a better algorithm, at least for the type of community detection that graphs like IMDB require. 
%
To design such an algorithm, we need to understand how communities such as those in the
IMDB network are formed. Towards that, we introduce a simple, but insightful generative model for community 
formation which we call the  {\em sequential community graph model}.  In this model, vertices (individuals) 
arrive sequentially and either join existing communities in the graph or form a new community.  Unlike the 
models in \cite{yang2013overlapping} and \cite{yangcommunity}, our model is a combinatorial model and 
does not have any (hyper-)parameters. 

The value of the model, in a sense, is in its ability to unearth communities hidden in a graph structure
using an appropriate algorithm. We show that, for a graph generated by the sequential community model, 
there exists an extremely simple algorithm which we call the leader-follower algorithm (LFA), that can find all the 
communities successfully (see Theorem \ref{thm:exact}). The key property that we identify to establish 
this result is that all sequential community graphs are chordal graphs, and the LFA algorithm is effectively
identifying the maximal cliques in this chordal graph. The LFA algorithm works for any graph and its running 
time is bounded by $O(|V|^2|E|)$ for a graph with vertex set $V$ and edge set $E$ (see Theorem \ref{thm:runtime}). 

While this running time is polynomial, it can be prohibitively expensive for very large graphs. To that
end, we propose a natural heuristic that simplifies the LFA algorithm, which we call {\em fast} leader-follower
algorithm (FLFA). It runs in time $O(|E| + |V| \log |V|)$ for graphs with vertex set $V$ and edge set $E$, 
i.e. FLFA is effectively linear in the input data size (see Theorem \ref{thm:runtimeFLFA}). We establish 
that the FLFA finds a specific subset of communities correctly for sequential community graphs 
(see Theorem \ref{thm:exactFLFA}). 

The purpose of the sequential community graph model was to identify an algorithm that can perform well for graphs like the IMDB graph, as mentioned earlier. We evaluate the performance of our algorithms on the IMDB
graph and find that both of them have F1 scores of 0.81, which is definitely better than the information lower bound of 0.5 (see Figure \ref{fig:IMDB}).
We evaluate the algorithm's performance for other datasets studied in the literature 
where {\em ground truth} communities are known. We find that for all such datasets, the LFA (and FLFA) 
outperform the representative known algorithms, namely, fast modularity optimization 
\cite{newman2006modularity} and statistical inference based methods (
CESNA \cite{yangcommunity} and BigClam \cite{yang2013overlapping}).
We note that the FLFA runs orders of magnitude faster than all the other algorithms. The 
precise results are described in  Section \ref{sec:results}). 

%
%

%
%
%
%


\medskip
\noindent
{\bf Related work. } 
There are multiple approaches for community detection.  Some are based on heuristics, such as modularity 
optimization \cite{newman2006modularity} and k-clique percolation \cite{palla2005uncovering}.  More recently,
there has been a lot of activity around developing statistical inference based algorithms for community detection
by positing probabilistic generative model for communities. This includes  the stochastic blockmodel and 
its variants \cite{airoldi2008mixed, handcock2007model,daudin2008mixture,newman2016estimating,karrer2011stochastic,
yang2013overlapping,yangcommunity, decelle2011asymptotic}.  One benefit of model based approaches is that they allow one to establish
theoretical performance guarantees \cite{krzakala2013spectral, hajek2016achieving,mossel2015consistency,abbe2016exact,abbe2015recovering}.

Many community detection methods can be difficult to implement exactly, but very often
efficient approximations have been found.  Modularity optimization
is well known to be an NP-hard problem \cite{kar72}, but a very efficient procedure
for modularity optimization is proposed in \cite{blondel2008fast}. Statistical inference based 
methods can suffer in terms of scaling with data size due to the complexity of the inference task, 
but clever approaches have helped overcome such challenges, for example \cite{decelle2011asymptotic, krzakala2013spectral, yang2013overlapping, yangcommunity}

\medskip
\noindent
{\bf Organization. }  The remainder of the paper is organized as follows.  
Section \ref{sec:score} introduces the F1 score for community detection algorithms 
as well as our result on a non-trivial lower bound for it. 
Section \ref{sec:axioms} presents the sequential community graph model. 
Section \ref{sec:algorithm} presents the leader follower algorithm (LFA) and its efficient
variant (FLFA). We establish their theoretical properties as well.  We present an empirical 
evaluation of our algorithms in Section \ref{sec:results} and conclude in Section 
\ref{sec:conclusion}. 

\section{The Score Function}\label{sec:score}
We are given an undirected graph $G = (V,E)$ where $V = \{v_1,\dots, v_n\}$ represents vertices and
$E \subset V \times V$ represents edges between them.  We refer to $G$ as an \emph{observation graph}
because it represents all observed interactions between the vertices.   
The observation graph is generated through some unobserved process by a set of latent
communities 
$\mathcal C = \curly{c_1,c_2,...,c_m}$, where $c_i\subseteq V$ for $i=1,2,...,m$.  
The community detection
problem is to use the observation graph $G$ to recover the latent communities $\mathcal C$.  

To assess the
accuracy of community detection algorithms, we define a 
score to compare sets of communities.  
For any two sets of communities $\mathcal C$ and
$\mathcal C'$ of an observation graph, we define their score as
\begin{align}
	d(\mathcal C,\mathcal C') & = \frac{1}{2}\paranth{s(\mathcal C, \mathcal C') + s(\mathcal C', \mathcal C)}\label{eq:f1}		
\end{align}
where we have defined $s(\mathcal C,\mathcal C')$ as

\begin{align}  
  s(\mathcal C,\mathcal C') & = {\frac{1}{|\mathcal C|}\sum_{c\in\mathcal C}
                                        \max_{c'\in\mathcal C'}\delta(c,c')}\label{eq:f2} 
\end{align}
and $\delta(c,c')$ is a similarity measure between two communities.  There are a variety
of similarity measures we can choose, but we will 
follow the approach of \cite{yangcommunity} and use the F1 score 
which is used commonly in binary classification.  For two communities
$c$ and $c'$, we define the precision $p=|c\bigcap c'|/|c'|$ and
the recall $r=|c\bigcap c'|/|c|$.  The F1 score is
given by the harmonic mean of $p$ and $r$: $\delta(c,c')=2pr/(p+r)$.  
For two identical community sets, the F1 score
is one and the minimum value of the F1 score is zero for two disjoint communities.

The quantity $s(\mathcal C, \mathcal C')$  finds the best match in $\mathcal C'$ 
for every community in $\mathcal C$.  It then calculates the average similarity
score of this matching.  
Note that multiple communities in $\mathcal C$ are allowed 
to match to the same community in $\mathcal C'$ 
to allow for the possibility that communities in $\mathcal C$ 
are subsets of the same community in $\mathcal C'$.
The overall score, $d(\mathcal C,\mathcal C')$ is simply the average of 
$s(\mathcal C,\mathcal C')$ and $s(\mathcal C',\mathcal C)$. 
To see why our score needs both $s(\mathcal C,\mathcal C')$ and 
$s(\mathcal C',\mathcal C)$, 
consider the case where $\mathcal C = \{a\}$ and 
$\mathcal C' = \{a, b, c, d, e, f\}$. 
If our score only accounted for $s(\mathcal C,\mathcal C')$, 
we would obtain a score of $1$, even though the communities 
are clearly quite different.  The quantity 
$s(\mathcal C',\mathcal C)=1/5$.
 Hence, we need to account for the 
both $s(\mathcal C,\mathcal C')$ and $s(\mathcal C',\mathcal C)$ 
to obtain an informative score for two sets of communities.

To understand what constitutes a good value of this score, 
we consider the set of communities which is the \emph{power set}
of the vertices.  The communities in this set are every
possible subset of $V$.  This is an extremely
trivial community set and provides no information about community structure.
We have the following result about the score of the power set communities
and any arbitrary set of communities.
\begin{lemma}\label{lem:powerset}
Let $\mathcal C$ be an arbitrary set of communities 
of a set of vertices $V$ and let the power set of $V$ 
be $\mathcal P$.  Then \[d(\mathcal C,\mathcal P)\geq 0.5.\]
\end{lemma}
This shows that the most uninformative community set will
score at least 0.5.  We will refer to this as the \emph{information lower bound}.  The output of a community
detection algorithm  
must produce a score greater that 0.5 in order
to be considered non-trivial.  This is an important result
because in previous works algorithms achieve scores below this
threshold, showing that no informative community structure has
been found \cite{yang2012community,yangcommunity,ruan2013efficient}
\begin{proof}
Every set in $\mathcal C$ matches exactly with one set in $\mathcal P$ and
will have an F1 score of one. Therefore $s(\mathcal C,\mathcal P)=1$ which
immediately leads to $d(\mathcal C,\mathcal P)\geq 0.5$.
\end{proof}

\section{Generating Communities: \\Sequential Community Graphs}\label{sec:axioms}
    \begin{figure}[t]
	\centering
		\includegraphics[scale = .9]{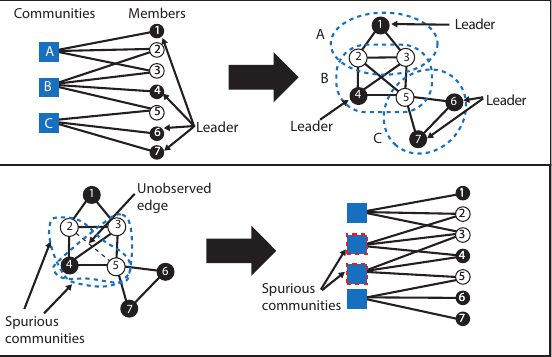}
		\caption{ A latent community graph (top left)  and its 
		corresponding observation graph (top right).  The leader vertices
		are colored black and the communities in the observation graph are
		circled with dashed lines.  (bottom) Example of spurious
		communities created by an unobserved edge in the observation graph.}  
	\label{fig:example}
\end{figure}

\subsection{Latent Community Graphs}

We assume that the observation graph $G$ is generated by an underlying
 latent or  unobserved community structure $\mathcal C$. 
To make this more precise, let
$\cG = (V, \cC, \cE)$ represent the bipartite \emph{latent community graph}, 
where one set of vertices is $V$ (the vertices
we observe in $G$) and the other set  
is ${\cC}= \{ c_1, \dots, c_m\}$ is the $m$ communities. 
The edges $\cE \subset V \times \cC$ are between these two sets, i.e. $\cG$ is bipartite. 
The edges of $\cE$ represent the membership of vertices of $V$ 
in communities of $\cC$: $(i,c) \in \cE$ if vertex $i$ belongs to community $c$. 
The observation graph $G = (V,E)$ is a {\em projection} 
of $\cG$: $(i,j) \in E$ if and only if vertices $i, j \in V$ 
share one or more communities in $\cG$, i.e. there exists 
$c \in \cC$ such that $(i,c), (j,c) \in \cE$.  
We illustrate these graphs in Figure \ref{fig:example}.

One property of the latent bipartite community graph is that the resulting
communities in the observation graph will be cliques.  The latent bipartite community graph
which explains the observation graph with the fewest number of communities
will be such that each community is a \emph{maximal clique}. 
Recall that a subset $c$ of vertices $V$
is called a clique if $\{(i,j): i\neq j \in c\} \subset E$; it is a 
maximal clique if there is no $c' \subset V$ such that $c \subset c'$ 
and $c'$ is a
clique as well. Note that
for any given $G$, it is feasible to find a $\cG$ so that $G$ becomes 
the corresponding projection of $\cG$, but the communities are not guaranteed
to be maximal cliques. Because such a set of communities may not be
informative, we focus on finding
communities which are maximal cliques.
 It is well known that the problem of finding maximal cliques 
in an arbitrary graph is computationally hard \cite{kar72}.  
The question of interest is {\em are there prevalent social phenomenon 
generating latent community graphs for which 
finding communities in the observation graph is easy?} 
To answer this question, we shall present the 
sequential community graph model next. 

A few remarks are in order before we present the model. 
First, our problem formulation as well as the latent 
community graph has been considered before 
 \cite{breiger1974duality,simmel2010conflict,feld1981focused,yang2012community}. 
Second, in practice there may be missing  edges or noise in an observation graph. 
Statistical inference based models allow for these missing 
edges via a probabilistic mapping from the latent
community graph to the observation graph \cite{yang2012community}.
In our situation, missing edges would cause true communities to 
no longer be cliques. For instance, if an edge is removed from a clique with $n$ vertices,
then it becomes the union of two overlapping cliques each with $n-1$ vertices.  Therefore, noise
or missing edges in an observation graph will result in the creation of spurious community vertices
in the corresponding latent community graph. We illustrate this
in Figure \ref{fig:example}.  For the purposes of establishing theoretical results, we shall
assume that $G$ is perfectly observed. However, as we shall see, our algorithms are robust
to noisy observations.

\subsection{Sequential Community Graphs}
Here we present a generative model for latent community graphs which
we call the sequential community graph model. 
This model should be treated as a {\em social hypothesis} applicable to a class of social
scenarios. In particular, this model is relevant to settings 
where individuals enter a social graph by either 
joining existing communities or creating their own. 
We now present the model in detail. 

\begin{figure}[t]
	\centering
		\includegraphics[scale = .68]{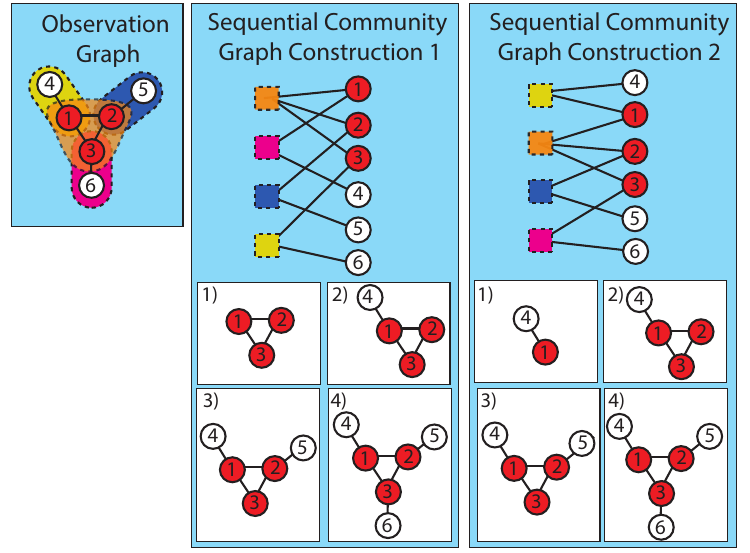}
		\caption{Illustration of two sequential community graph constructions
		(and the sequential construction of the corresponding observation graph)
		resulting in the same observation graph.  In each construction 
		the member vertices are listed in order of addition to the graph with the newest vertex
		at the bottom.}  
	\label{fig:sequence}
\end{figure}

Let $\cG_n=(V_n,\cC_n,\cE_n)$ denote a 
sequential community graph with $n$ observed vertices, 
i.e. $|V_n| = n$.  
This graph is generated sequentially as follows. 
Initially, $n = 1$ and $V_1 = \{v_1\}$, $\cC_1 = \{c_1\}$ and $\cE_1 = \{(1,c_1)\}$. 
Given $\cG_i$, $\cG_{i+1}$ is generated by adding vertex $v_{i+1}$ to $V_{i}$, i.e. 
$V_{i+1} = V_i \cup \{v_{i+1}\} = \{v_1,\dots, v_{i+1}\}$.  
For $\cC_{i+1}$ and $\cE_{i+1}$, one of the two choices listed below
is exercised arbitrarily: 
\begin{enumerate}

\item[]{\em Choice 1.} Choose a single community, $c \in \cC_i$; 
add edge $(v_{i+1}, c)$ to $\cE_i$ to obtain $\cE_{i+1}$ and set $\cC_{i+1} = \cC_i$.

\item[]{\em Choice 2.} Add a new community vertex $c'$ to $\cC_i$ to obtain $\cC_{i+1}$
and add a new edge $(v_{i+1},c')$ to $\cE_i$ to obtain $\cE_{i+1}$.  Then 
select any one other community vertex $c \in \cC_{i}$.  Let 
$V_c = \{v \in V: (v,c) \in \cE_i\}$ be the neighbors of $c$ and select
an arbitrary proper subset
$V'_c\subset V_c$ ($V'_c$ can also be the empty set). 
Add edges $\{(v, c'): v \in V'_c\}$ to $\cE_{i+1}$. 

\end{enumerate}
In a sequential community graph $\cG_n$ 
there can be a maximum of $n$ community vertices because a new community
vertex can only be generated by a new observation vertex.
Also note that the construction
of a sequential community graph is not unique.  
There can be multiple sequences of vertices
that produce a given sequential community graph. We illustrate
this with an example in Figure \ref{fig:sequence}.

The sequential community graph model corresponds to social 
phenomena where new members join a social network
by either joining an existing community
or generating a new community from a subset of an existing community.
Thus, new communities are only 
created when new members join the graph.  This is not an unreasonable
assumption. Consider for example the graph
formed by the friends of an individual in an online social network.  Communities 
are the mechanism by which   people become
friends with this individual. Either the friendship
is formed from an existing community, or a brand new community is formed.
The sequential community graph model assumes that a new friendship can only occur
from a single community and that new communities can only include
members of a single existing community.  While this restricts the possible community
structures, it does allow for efficient and exact recovery of communities.


The sequential community graph model motivates us to
divide the vertices in any observation graph 
into two types.  Recall that in this model, a community
is a maximal clique.  The vertices can be divided into
those that belong
to single and multiple communities/maximal cliques.
  We define these vertex types as follows.
\begin{dfn}
A vertex in an observation graph is 
a \textbf{leader} if it
belongs to only one maximal clique. Otherwise it is 
a \textbf{follower}. 
\end{dfn} 
We call vertices which belong to a single community 
leaders because they are the individuals in our model 
whose ``loyalties'' lie in a single community.
In graph theoretic terms, they are known as \emph{simplicial vertices},
which are vertices whose neighbors induce a subgraph that is a clique \cite{wainwright2015graphical}.
For example,  in an individual's online social network, leaders are  
the people the individual only knows through a single community. 
Everyone else is naturally deemed to be a follower because
the individual knows them through multiple social contexts and so they do
not uniquely correspond to a single community.  We illustrate
the notion of leaders and followers in the example in Figure \ref{fig:example}. 

The construction of a \scg naturally incorporates
our notions of leaders and followers.  A new community
can only be generated by a leader.  
Followers belong to multiple communities and do not truly
give a community its identity.  
As a sequential community graph evolves,
the roles of vertices can change.  In particular,
leaders can become followers if they join
communities that new leaders have created.

The sequential community graph has many important
properties that facilitate fast community detection.
One important property is that it has a perfect
elimination ordering, which we define now.

\begin{dfn}[\cite{wainwright2015graphical}]
Consider a graph $G=(V,E)$. Let $(v_1,v_2,...,v_n)$
be a \textbf{perfect elimination order} of the vertices in $V$.
Then for each vertex $v_i$, the subgraph induced
by $v_i$ and its neighbors in $(v_{i+1},v_{i+2},...,v_n)$
form a clique.
\end{dfn} 
For sequential community graphs, we have the following
result.
\begin{lemma}\label{lem:poe}
Let the vertex sequence for a sequential community graph be $(v_1,v_2,...,v_n)$.
Then a perfect elimination order for the graph is $(v_n,v_{n-1},...,v_1)$.
\end{lemma}
Here we see that the reverse order in which vertices
join the graph is a perfect elimination order. 
\begin{proof}
We prove the result by establishing a contradiction.
Assume that the sequence $v_n,v_{n-1},...,v_1$ is not
a perfect elimination order.  Then there must be some
vertex $v_i$ such that its neighbors in $(v_{i-1},v_{i-2},...v_1)$
do not form a clique.  However, by the rules of construction for
a sequential community graph, when $v_i$ joins the graph,
it either joined one existing community or formed a new community with vertices from
one previous community.  Either way, $v_i$ and its neighbors among the vertices
that joined before it form a clique, which is a contradiction.
\end{proof}

The existence of a perfect elimination ordering for a sequential
community graph puts it in a special category of graphs, as shown
by the following result.

\begin{thm}\label{thm:chordal}
A sequential community graph is a chordal graph.
\end{thm}
\begin{proof}
By Lemma \ref{lem:poe}, a sequential community graph
has a perfect elimination order.  By definition, a graph is chordal if and only if 
it has a perfect elimination order \cite{wainwright2015graphical}.
\end{proof}

Because sequential community graphs are chordal,
they possess important properties
 which allow us to efficiently 
recover all of their communities.  We now present
some of these properties.
%

\begin{dfn}[\cite{wainwright2015graphical}]\label{dfn:recsimp}
A graph is recursively simplicial if it contains a simplicial vertex $v$ and when $v$
is removed the subgraph that remains is recursively simplicial.\end{dfn}
\begin{prop}[\cite{wainwright2015graphical}]\label{prop:deletion}
Chordal graphs are recursively simplicial.
\end{prop}

This property shows that after removing the leaders of a community (which
are simplicial vertices) from
the observation graph, the remaining graph will still be a sequential 
community graph.  This recursive simplicial property
is the key idea behind our community detection algorithms
in Section \ref{sec:algorithm}.


\section{Leader-Follower Algorithms}\label{sec:algorithm}
We use the notion of followers and leaders and the properties of
\scgs discussed in Section \ref{sec:axioms}
to develop two community detection algorithms:  the fast leader-follower algorithm (FLFA)
and the  leader-follower
algorithm (LFA).  Both algorithms are able to
detect overlapping communities.   The FLFA is a simple procedure
which can detect communities very quickly.  The LFA
is an iterative procedure which involves running the 
FLFA as a subroutine and then
removing certain vertices from the observation graph.  
The LFA can find more communities than FLFA because 
it is applied iteratively to a transformed observation
graph.   However, we will see in practice that both 
algorithms have very similar performance
in terms of accuracy, but the FLFA has a strong 
advantage in terms of speed.

\subsection{Fast Leader-Follower Algorithm}\label{sec:flfa}
      \begin{figure}[t]
	\centering
		\includegraphics[scale = 0.57]{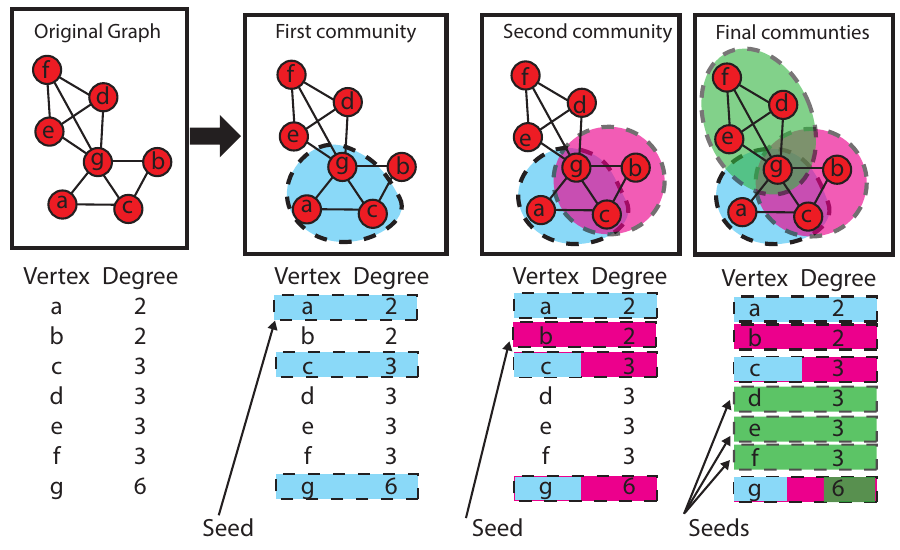}
		\caption{Application of FLFA to a graph with three communities.  (top) The figures show
		 each new community that is detected.  (bottom) The list of degree ordered vertices has 
		(multi)colored
		rectangles showing the (possible multiple) community 
		membership of the vertices as new communities are detected.  The seeds of
		each new community are indicated in the vertex lists.}  
	\label{fig:flfa}
\end{figure}

The key to the FLFA is the fact that each community in a \scg 
can be identified by finding its leaders. Since the leaders of a community
only belong to one community, the neighbors of the leaders will constitute 
the entire community. Thus, finding the leaders associated with a community 
allows us to find all the members of the community. 

To find leaders, 
FLFA makes use of the fact that the degree of a leader must 
be less than or equal to the degree of its neighbors, due to 
the fact that a leader only has connections to vertices within 
a single community.  Thus, to find leaders, 
FLFA simply attempts to find vertices whose degree is less
than or equal to their neighbors.  Once leaders are found, 
their neighbors determine the underlying community structure 
in the graph.

FLFA uses the following approach to find leaders 
in a graph quickly. It orders the vertices from lowest to highest
degree.  Since leaders have a  lower degree than followers, 
leaders will naturally appear earlier in the list. 
It then  iterates through the list and finds the first 
vertex that has not been marked as visited yet. 
It marks the vertex and all of its neighbors as visited.
The vertex and all of its neighbors are then placed into a community. 
We refer to the minimal degree vertices in a community
found by the FLFA as the \emph{seeds} of the community.  Note
that seeds are not necessarily leaders as we have define them (i.e. simplicial vertices).  
Rather, they represent an approximation 
for what the leaders may be in the observation graph. 
As such, the communities that are found are not necessarily cliques. 

The FLFA is able to find communities in the graph extremely quickly using just a single pass
through the vertices. Moreover, it is also succinct and simple 
in its description and implementation. Lastly, as we 
shall see in the results section, it still is able to 
find communities with a relatively high accuracy, despite 
taking a fraction of the time of other algorithms. 
We illustrate the application
of the FLFA to an example graph in Figure \ref{fig:flfa}.
The steps of the FLFA are specified below.
\newline 
\begin{algorithmic}
	\Procedure {FLFA} {$G$}
		\State $\cC \gets \{\}$, $Visited \gets \{\}$
		\State $L \gets$ list of vertices of $G$ sorted by ascending degree
			
		 \For {$ 1\leq i \leq \text{length}(L)$}
			\State $v \gets L [i]$
			\If {$v\notin Visited$}
				\State $c \gets v \cup \text{Neighbors}(v)$
				\State $\cC \gets \cC\bigcup c$  
				\State $Visited \gets Visited\bigcup c$
			\EndIf
		\EndFor
		\State return $\cC$
	\EndProcedure
\end{algorithmic}

%

\subsection{Leader-Follower Algorithm}\label{sec:ilfa}

For some graphs the FLFA is not able to find all the communities. 
During the construction of the sequential community graph, this occurs when 
new leaders enter the graph and cause leaders of a previous community 
to become followers. The key to discovering a leader for these 
hidden communities is to remove the vertices that caused the leaders 
of the given community to become followers. 
This motivates what we call the  
\emph{leader-follower algorithm} (LFA) for 
community detection.  This algorithm is designed to detect
communities which cannot be found by the FLFA. 

At each iteration of the LFA, we choose a simplicial vertex
in the graph and form a community from it and its neighbors.  If
the community is a clique and not a subset of a previous community,
we include in the set of detected communities.  We then delete
the vertex from the graph.  This iteration is repeated until the
graph is empty.  With these steps, we obtain a robust algorithm that, as we will see, 
can exactly discover all the communities in any sequential community graph.
The steps of the LFA are specified below.

\smallskip
\begin{algorithmic}
	\Procedure {LFA} {$G$}
		\State $\cC \gets \{\}$
		\While {$G$ \text{is not empty and has a simplicial vertex}}
			\State $v =$\text{simplicial vertex in~}$G$
			\State $c \gets v \cup \text{Neighbors}(v)$
			\If {$c$ is a clique and $c$ is not a subset of any $c'\in\cC$}
			  \State 	$\cC\gets c\bigcup \cC$
			\EndIf
			\State $G \gets G-\curly{v}$
		\EndWhile
		\State return $\cC$
	\EndProcedure
\end{algorithmic}


 \section{Performance Guarantees}\label{sec:theory}

We will next establish theoretical performance guarantees for the LFA and FLFA.  
The main results presented here concern the performance of the algorithms in terms of  accuracy and speed.

\subsection{Accuracy} 
Recall that in the observation graph for a
latent community graph, the communities are maximal cliques.
This makes community detection for this model equivalent to finding
maximal cliques.  
The LFA and FLFA were designed to find maximal cliques
 and their performance
is strongest in graphs where communities take this form, such as
 sequential community graphs.

We first present our result for the FLFA.
There are examples of \scgs
 where the FLFA cannot find all communities.  Therefore, FLFA cannot
detect communities on all sequential community graphs.
However, there is a subclass of \scgs where the FLFA
will detect all communities.  Our result is as follows.
\begin{thm}\label{thm:exactFLFA}
	Let $G = (V,E)$ be the observation 
	graph of a sequential community graph. The output of the FLFA applied 
	to $G$ will contain every maximal clique of $G$ that has a leader.
		\end{thm}
\begin{proof}
Consider an observation graph $G=(V,E)$ and let $c\subseteq V$
be a set of vertices forming a maximal clique with at least one
leader.  Let one of these leaders be $l$.  Because $l$ is a leader,
all of its neighbors are in $c$ and it has degree less than or
equal to all of its neighbors.  
In the degree sorted list used in the FLFA, $l$ and all of its 
neighbors of equal degree will occur before the non-leaders in $c$.  
We assume without loss of generality that $l$ occurs in the 
degree sorted list before all other vertices in $c$.  
$l$ is not assigned to any community created by vertices that occur
before it in the degree sorted list because it does not neighbor any of them.
It is is the first vertex in $c$ that the FLFA identifies as a seed.  
The FLFA forms a community corresponding to $l$ and all of its neighbors, which
is equivalent to $c$.  Therefore, the FLFA output will contain $c$.  
Because this result holds for any maximal clique in $G$ with at least one leader,
the FLFA output will contain all such maximal cliques.
\end{proof}

This result shows that the FLFA has exact 
detection on the subclass of sequential
community graphs where each community has a leader,
but in many \scgs leaders become followers as the graph
evolves.  To achieve correct detection
for the general class of \scgs 
we require the LFA.
Our formal result is the following.
\begin{thm}\label{thm:exact}
	Let $G = (V,E)$ be the observation graph of a sequential community graph. 
	The output of LFA applied to $G$ will be the exact set of maximal cliques in $G$.
	\end{thm}
\begin{proof}
For a sequential community graph $\cG$, we define its communities as $\cC$ 
and its observation graph as $G$. 
Recall that
because $G$ is the observation graph of a sequential community
graph, every member of $\cC$ corresponds to a maximal clique
in $G$.
We define the output of the LFA applied to $G$ as $\cC_{LFA}$.
To prove Theorem \ref{thm:exact} we show that 
$\cC = \cC_{LFA}$.  

\textbf{Every $c\in\cC$  is in $\cC_{LFA}$.}  First we consider
$c\in\cC$ which has at least one leader $l$.  Because $l$ is a simplicial vertex, its non-simplicial neighbors will never be deleted before it. 
At some iteration, $l$ (or its simplicial neighbor if exists) will be chosen to form the community with all its neighbors and be placed in 
$\cC_{LFA}$.

Now consider $c\in\cC$ which does not have a leader.  To establish that this community will be found by the LFA, we first construct a 
clique tree for $G$. We define the clique tree $G_C = (\cC,E_C)$ with $(c_1,c_2)\in E_C$ if
$c_1\bigcap c_2 \neq \emptyset$.  That is, each community is a vertex and there is an edge between two vertices if their corresponding 
communities have a non-empty intersection.  In the construction of a sequential community graph
we either add no new communities or add a single community which is joined
by members of at most a single previous community.  In the clique tree, this means
that each community has at most one parent, which guarantees that it is a tree (we assume without loss of generality that $G_C$ is connected).  

Each leaf in $G_C$ must have at least one leader, otherwise it would be a subset of its parent.  Eventually an iteration of the LFA will find one of these leaf communities and remove one of their leaders.  When all leaders are deleted, the leaf is removed from $G_C$, because without its leaders it is a subset of its parent and is no longer a maximal clique in the updated observation graph.  Because we assumed $c$ has no leaders, it is not detected until it becomes a leaf in $G_C$.    As the leaves are removed in the clique tree, at some iteration $c$ will become a leaf and possess a leader in the corresponding observation graph.  None of the vertices in $c$ will be deleted until  $c$ contains a simplicial vertex.  At this iteration when $c$ is a leaf in the clique tree and has a minimal degree vertex, it is detected and placed in $\cC_{LFA}$.
\newline
\newline
\textbf{Every $c\in\cC_{LFA}$ is in $\cC$.}  Recall from Proposition \ref{prop:deletion} that $G$ is a recursively simplicial graph.  This means that when a leader is deleted, the remaining graph will have at least one leader.  Each iteration will find a community with a leader.  Furthermore, this community is a maximal clique in the corresponding observation graph.   Therefore, each iteration is guaranteed to find a maximal clique with at least one leader in the current observation graph.  

Let $c$ be one of the communities found in an iteration of the LFA.  One possibility is that $c$ is a maximal clique of the original observation graph, so $c\in\cC$.  The other possibility is that $c$ is a subset of a maximal clique $c'\in\cC$. In the latter case, $c$ is only a subset of $c'$ because some vertices in $c'$ were deleted in a previous iteration.  But this can only happen if these vertices were leaders, which means $c'$ has already been detected by the LFA, so we have $c'\in \cC_{LFA}$.  
\end{proof}



\subsection{Runtime}  
We now analyze the runtime of the FLFA and LFA.  
Our first result concerns the runtime of the FLFA.
 \begin{thm}\label{thm:runtimeFLFA}
        For an input graph $G(V,E)$, the FLFA will terminate in $O(|E| +|V|\log(|V|))$ time. 
 \end{thm} 
As can be seen, the FLFA is very fast with a runtime that is linear in the graph size.
  
  \begin{proof}
	 The first step of the FLFA is to calculate the degree of each vertex
	and sort the vertices by degree.  Calculating the degree involves counting
	every edge in the graph at most twice which takes $O(|E|)$ time.
	Sorting the $|V|$ vertices can be done in $O(|V|\log(|V|)$ time.  The second step
	is to go through the degree sorted list and assign each unvisited vertex and its
	neighbors to a community.  This can be done in $O(|E|)$ time.  Combining these steps,
	we find that the a total runtime of the FLFA is $O(|E|+|V|\log(|V|))$.
\end{proof}
We have the following result for the LFA runtime. 
 \begin{thm}\label{thm:runtime}
        For an input graph $G(V,E)$, the LFA will terminate in 
				$O(|V|^2|E|)$ time. 
 \end{thm} 
 The runtime of the LFA is determined by the number of iterations it requires to
terminate.  While the worst case bound in Theorem \ref{thm:runtime} can be potentially
large, we will see in Section \ref{sec:results} that in practice FLFA and LFA
have very similar runtimes on large graphs because not many iterations of FLFA are
needed.

\begin{proof}
Each iteration of the LFA involves finding a simplicial vertex, checking
if it and its neighbors form a community that is a clique and not a subset of a previous community, and then deleting this vertex from the observation graph.
Finding a simplicial vertex takes  $O(|E|)$ operations.
Checking if a single community is a clique and a subset of a previous
community will require at most $|E|$ 
operations, and there cannot be more than $|V|$ communities. Using this, we find that each iteration of the LFA will require
 $O(|V||E|)$
steps.  For a graph of $|V|$ vertices, the maximum number of iterations is $|V|$.  Therefore,
the worst case runtime of the LFA will be $O(|V|^2|E|)$.
\end{proof}

\section{Empirical Evaluation}\label{sec:results}
We now compare the performance of the LFA and FLFA to other
state of the art community detection algorithms on several
real graphs. We compare the performance of the algorithms 
in terms of accuracy and speed
on graphs with known ground truth communities.  
The algorithms we compare against include the method 
for fast modularity optimization \cite{blondel2008fast}
and methods based on probabilistic generative models: 
CESNA \cite{yangcommunity} and BigClam \cite{yang2013overlapping}.

\subsection{Data Description }\label{sec:data}
Our dataset consist of several graphs for which 
we have accurate ground truth communities.  
We describe these graphs below.  All properties
of the graphs are shown in Table \ref{table:graphs}.

\begin{table}[ht]
  \centering
    \begin{tabular}{|l|r|r|r|r|}
    \hline
     Graph& $|V|$& $|E|$& $|\mathcal C|$\\\hline
     Prime number graph& 999 & 195,309 & 168 \\\hline
      Culture show 2010& 153 &1802 & 13 \\\hline
       Culture show 2011& 138 &3626 & 10\\\hline
       Les Miserables & 71  &244  & 80\\\hline
     IMDB & 382,219  &15,038,083  & 127,823\\\hline
    \end{tabular}
  \caption{Graph properties:  Number of member vertices $|V|$, number of edges $|E|$, and
  number of communities $|\mathcal C|$.}
  \label{table:graphs}
\end{table}


\medskip
\noindent{\bf Prime Number Graph.} 
 In a prime number graph with $N$ vertices,
the integers from $2$ to $N+1 > 2$ are vertices, 
edges between two integers indicate that they share 
a prime number as a common factor (e.g. $14$ and $21$
have an edge since they have $7$ as a common factor), 
and a community corresponds to a prime number in
the sense that it is a collection of integers all of 
which have a given prime as their factor (e.g. all integers that
contain $7$ as a factor). 

We use a prime number
graph whose vertex set
is the integers from $2$ to $1,000$.  
The number of ground truth communities 
is 168, which is the number of prime numbers less than 1,000.
There is great heterogeneity in the community sizes, with some communities
constituting half of the vertices, while others being isolated vertices. 

\medskip
\noindent{\bf Culture Show Graphs.} 
The culture show 2010 and 2011 graphs represent performances 
	from a college culture show at MIT in 2010 and 2011.  
	The vertices are performers and the edges indicate 
	whether or not two performers were in the same performance. 
	Each performance is a separate ground truth community in this graph.
	
\medskip
\noindent{\bf Les Miserables Graph.}
The Les Miserables graph captures the social interactions of the characters
in the novel Les Miserables.  The vertices are characters from the novel 
and an edge is placed between two characters 
if they appear in the same chapter of the novel.  
Each chapter corresponds to a
 separate ground truth community in this graph.

\medskip
\noindent{\bf Internet Movie Database (IMDB) Graph.}

The IMDB graph consists of actors in movies \cite{ref:imdb}.  
Each vertex is an actor and an edge is placed between two actors if they performed
in the same movie. Each ground truth community 
consists of actors who were all in the same movie. 
This graph is very large, with 382,219 vertices (actors) 
and 127,823 communities (movies).  We will use this graph
to demonstrate that our algorithms scale to larger graphs while also
maintaining good accuracy.


\subsection{Experimental Results}

\begin{figure*}[t]
	\centering
		\includegraphics[scale = .51]{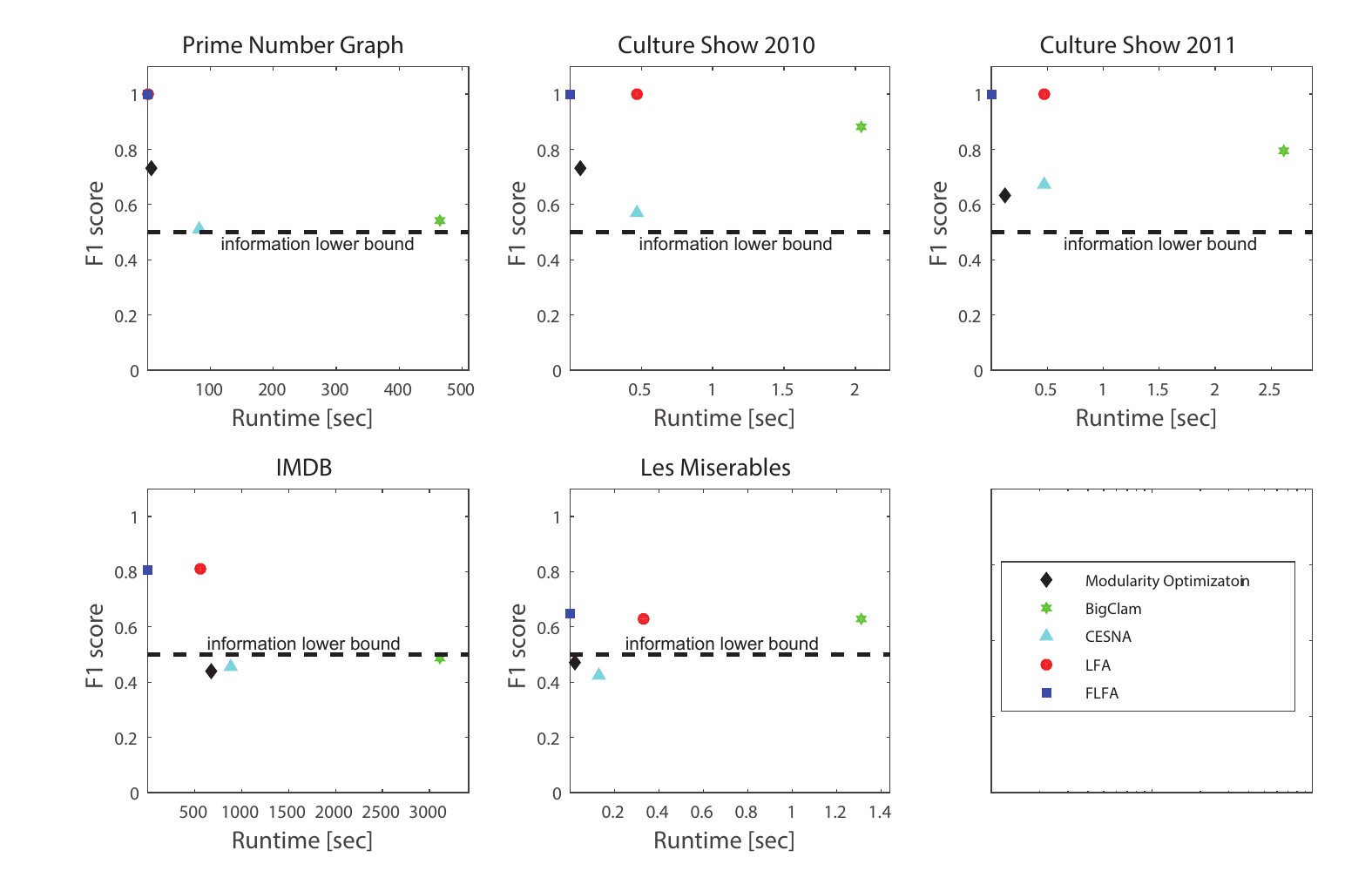}
		\caption{Plot of F1 score  versus
			the runtime of each algorithm.
			The dashed line indicates the 0.5 information lower bound.}  
	\label{fig:f1}
\end{figure*}
\begin{table*}[ht]
	\centering
		\begin{tabular}{|l|c|c|c|c|c|c|c|}
		\hline
		  Algorithm&\shortstack{Prime \\number\\ graph}& \shortstack{Culture\\ show\\ 2010}& \shortstack{Culture\\ show\\ 2011}& Les Miserables&IMDB \\\hline
		 Ground-truth &168& 13   & 10          & 80& 127,823 \\\hline
		\shortstack{Modularity\\ optimization} &105& 7      & 5       &5   & 2198\\\hline
		BigClam &56& 45      & 38          & 8& 100 \\\hline
		CESNA &2& 2        & 2          & 2& 2  \\\hline
		LFA &\textbf{168}& \textbf{13} &\textbf{10} &\textbf{34} & \textbf{61,059}\\\hline
		FLFA &\textbf{168}& \textbf{13} &\textbf{10} &30 & 60,876  \\\hline
		\end{tabular}
	\caption{Number of communities produced by each algorithm.
	The most accurate algorithms in terms of community number 
	for each graph are highlighted in bold.}
	\label{table:communities}
\end{table*}
%

We compare the performance of  FLFA and LFA to other algorithms
on these graphs. Figure \ref{fig:f1} shows the resulting
F1 score (equation \eqref{eq:f1}) and runtimes of each algorithm.

\begin{figure*}[h]
	\centering
		\includegraphics[scale = .65]{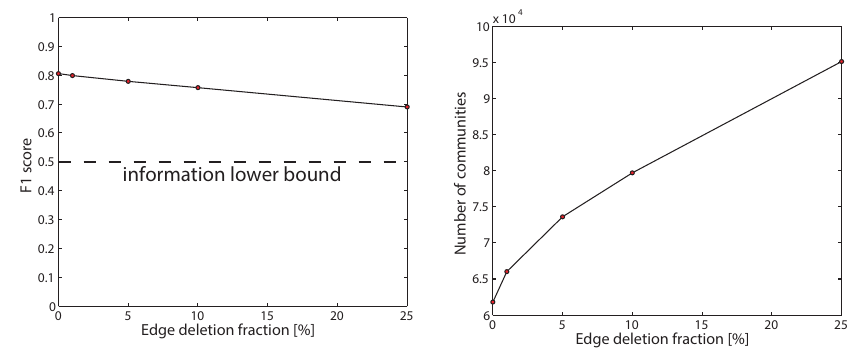}
		\caption{Plot of F1 score (top) and number of communities
			found (bottom) by FLFA on IMDB graph versus edge deletion fraction.
			The dashed line indicates the 0.5 information lower bound.}  
	\label{fig:robust}
\end{figure*}

\subsubsection{Accuracy}
Figure \ref{fig:f1} shows that FLFA and LFA perform well in terms of
accuracy on these graphs, 
consistently obtaining the highest scores. 
In the prime number graph, FLFA and LFA detect 
all communities exactly, obtaining a score of $1$, 
outperforming the next highest performing algorithm by 
$23\%$. 
Similarly, in the culture show graphs, 
FLFA and LFA again outperform the other algorithms. 
On both culture show graphs, FLFA and LFA both achieve a perfect score of $1$.
The next best algorithm achieves a score of $0.88$ on culture show 2010
and a score of $0.79$ on culture show 2011. 

In the Les Miserables graph, the LFA and FLFA have the best score
of $0.65$.  While this is not the perfect score we had on the prime number
and culture show graphs, it is greater than
the information lower bound of 0.5 given by Lemma \ref{lem:powerset}.  
Finally, on the IMDB graph, FLFA and LFA once again 
detect communities extremely well. As can be seen in the table, 
FLFA and LFA achieve a score of $0.81$.  The other algorithms
are not able to even cross the information lower bound.

In addition to having the best scores, the LFA and FLFA
also are the most accurate in terms of number of communities
found, as seen in Table \ref{table:communities}.  In some
instances, such as the IMDB graph, they are the only algorithms
that come within the same order of magnitude
of the number of ground-truth communities.

\subsubsection{Runtime}

Not only are FLFA and LFA the most accurate 
algorithms on these datasets, but they are also 
the fastest. As shown in Figure \ref{fig:f1}, 
FLFA and LFA consistently perform orders of magnitude 
faster than alternate methods. In particular, the FLFA 
is able to run much faster than the other algorithms.

What is even more striking is the fact that the FLFA
achieves this incredible speed with
without sacrificing much in accuracy.  The FLFA is the only algorithm
which simultaneously has a very fast runtime
and high accuracy. This is most evident on
the large IMDB graph, where the FLFA has
an F1 score of 0.81, which is nearly double that
of the other algorithms, yet has a runtime
under one second, which is almost \emph{three orders
of magnitude} faster than the other algorithms.
\subsubsection{Robustness}

Very often we will have missing data in an observation
graph.  We would
like to know how robust our community detection algorithms
to this type of noisy observation.  To check robustness, we perform
the following experiment.  We randomly remove different fractions
of edges from the IMDB graph and apply the FLFA.
The results are
shown in Figure \ref{fig:robust}.  As more
edges are deleted, the F1 score decreases, but
not substantially.  With 25\% of the edges removed,
the score decreases by only 12.5\%.  This
shows that the FLFA's performance
is not significantly degraded by missing data.

We saw earlier that missing data would result in
spurious communities being found.  From Figure \ref{fig:robust}
we see that this is indeed the case.  With
full observation, 61,876 communities were found by
the FLFA.  At 25\% edge deletion, this number 
grows by 50\%.  These spurious communities 
generally have strong overlap with the communities
found with no missing data, so even though
they are numerous, their impact on the
score is not as strong.

\section{Conclusion}\label{sec:conclusion}
A lower bound on the F1
community score function was established in order to
assess the non-triviality of the output of any community detection algorithm.
This is important because many algorithms were found to produce
community scores which were below this lower bound, thus bringing
into question the validity of their community outputs.

We presented the leader-follower and fast leader-follower algorithms (LFA and FLFA) for 
fast and accurate overlapping community detection.  We proposed a new generative
model for community formation in social networks based on very
natural social interactions.  We proved that the LFA
and FLFA were able to accurately learn the community structure of these models.
This provided a theoretical guarantee to the performance of the algorithms.

Experiments on graphs with ground truth communities showed that
the LFA and FLFA perform better than many state of the art 
algorithms which very often have community scores below
the trivial lower bound.  
The FLFA was found to be almost three orders of magnitude faster than
other algorithms while simultaneously maintaining high community detection accuracy.  
	This suggests that it can be used to 
	perform accurate, real-time community detection
	on extremely large graphs.
\bibliographystyle{abbrv}
\bibliography{lfa2014}

\end{document}